\documentclass[sigconf, nonacm]{acmart}
\usepackage{algorithm}
\usepackage{algorithmic}
\usepackage{multirow}
\AtBeginDocument{%
  }

\begin{document}

\title{DeeAD: Dynamic Early Exit of Vision-Language Action for Efficient Autonomous Driving}
\author{Haibo Hu}
\email{haibohu2-c@my.cityu.edu.hk}
\affiliation{%
  \institution{City University of Hongkong}
  \city{Hong Kong}
  \country{China}
}
\author{Lianming Huang}
\email{lmhuang8-c@my.cityu.edu.hk}
\affiliation{%
  \institution{City University of Hongkong}
    \city{Hong Kong}
  \country{China}
}
\author{Nan Guan}
\email{nanguan@cityu.edu.hk}
\affiliation{%
  \institution{City University of Hongkong}
    \city{Hong Kong}
  \country{China}
}
\author{Chun Jason Xue}
\email{jason.xue@mbzuai.ac.ae}
\affiliation{%
  \institution{Mohamed bin Zayed University of Artificial Intelligence}
    \city{Abu Dhabi}
  \country{The United Arab Emirates}
}

\renewcommand{\shortauthors}{Trovato et al.}

\begin{abstract}
Vision-Language Action (VLA) models unify perception, reasoning, and trajectory generation for autonomous driving, but suffer from significant inference latency due to deep transformer stacks. We present \textbf{DeeAD}, a training-free, action-guided early-exit framework that accelerates VLA planning by evaluating the physical feasibility of intermediate trajectories. Instead of relying on confidence scores, DeeAD terminates inference when predicted trajectories align with lightweight planning priors (e.g., Navigation or Low-precision Planning) within a tolerable deviation ($\leq$ 2m). To improve efficiency, we introduce a multi-hop controller that adaptively skips redundant layers based on the change rate of scores. DeeAD integrates into existing VLA models, such as ORION, without requiring retraining. Experiments on the Bench2Drive benchmark demonstrate up to 28\% transformer-layer sparsity and 29\% latency reduction, while preserving planning quality and safety.
\end{abstract}



\maketitle
\section{Introduction}
Recent progress in Vision-Language Models (VLMs)~\cite{li2023blip, liu2023visual, alayrac2022flamingo} has opened new possibilities for autonomous driving, enabling richer scene understanding and interpretable decision-making. 
Building on this momentum, recent systems such as ORION~\cite{orion,hao2025driveaction,kim2024openvla} extend large-model reasoning into the action domain, resulting in VLA frameworks that unify perception, language-based reasoning, and trajectory generation into a single end-to-end (E2E) architecture.

\begin{figure}[t]
  \centering
  \includegraphics[width=0.5\textwidth,trim=0cm 3cm 0cm 0cm, clip]{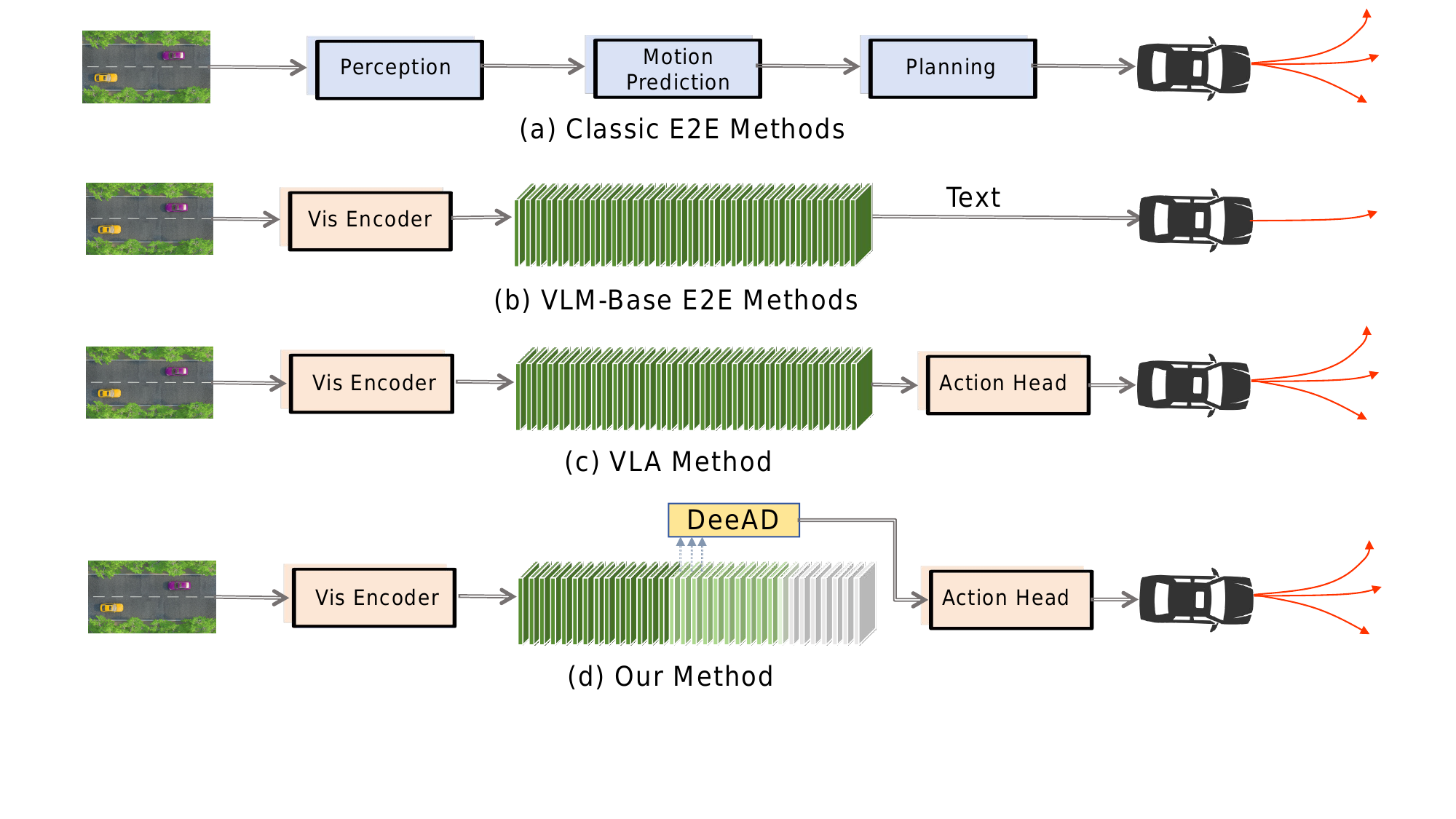}
  \vspace{-10pt}
  \caption{
Comparison of autonomous driving paradigms: 
(a) classic E2E, (b) VLM-based, (c) VLA, and (d) our DeeAD with Action-Guided Early Exit for efficient, physically consistent inference.
    }
  \vspace{-10pt} 
  \label{fig:vs}
\end{figure}

Fig.~\ref{fig:vs} shows the evolution of autonomous driving paradigms, ranging from classic modular pipelines~\cite{unidad2023, jiang2023vad}, to VLM-enhanced perception models~\cite{tian2024drivevlm, hwang2024emma}, and finally to full-stack VLA systems~\cite{orion}. 
While these unified models demonstrate impressive capabilities in simulation and closed-loop driving, they also impose significant computational overhead. 
A single inference pass typically involves dozens of transformer blocks, leading to hundreds of milliseconds of latency and heavy GPU utilization. 
This cost scales with model depth and input length, creating a fundamental barrier to real-time deployment on embedded platforms~\cite{deebert2020, deervla2024, layerskip2024, du2023accelerating}.

Our empirical analysis on the Bench2Drive benchmark~\cite{bench2drive2024} reveals that trajectory refinement across layers often saturates: 
intermediate layers can already generate safe and physically plausible trajectories that deviate only slightly from the final output. 
In some cases, these mid-layer plans align better with the intended navigation path than the final prediction. 
This mirrors real-world driving behaviors, where slight deviations from the optimal path are tolerated as long as the motion remains safe and consistent~\cite{modeseq2025}.

This observation motivates the need for an adaptive inference mechanism that dynamically terminates reasoning once a good-enough action is found. 
However, existing early-exit methods, typically based on feature-space confidence scores~\cite{deebert2020, cebert2024, pabee2023} or learned exit policies~\cite{layerskip2024, deervla2024, wu2020intermittent}, suffer from domain dependency and lack physical grounding, 
making them unstable under distribution shifts~\cite{bench2drive2024, hiddenbias2024}. 
This is particularly risky for autonomous driving, where inference uncertainty can directly affect safety.

To address this, we introduce \textbf{DeeAD}, a lightweight, plug-and-play early-exit framework tailored for VLA models. The core idea is intuitive: rather than waiting for the final decoding layer, the model should dynamically terminate inference once the predicted trajectory already falls within an acceptable deviation from a navigation reference. When intermediate predictions align closely, typically within a 2\,m tolerance, with route priors such as map navigation or low-precision planning, continued deep-layer reasoning brings diminishing returns. This follows a natural principle of human driving, where slight spatial deviations are permissible as long as motion remains safe and goal-consistent.

Turning this idea into a practical inference mechanism, however, introduces several challenges. First, standard VLA architectures only expose the final planning output, making it impossible to assess intermediate behaviors. To address this, we equip the decoder with an \emph{Early Exit Action Head} capable of extracting partial trajectories from any layer without retraining. Second, early termination requires a reliable measure of action-space consistency. We therefore introduce a lightweight \emph{Dissimilarity Estimation} component that evaluates the spatial deviation between predicted and reference trajectories, using metrics such as L2 displacement to quantify whether an intermediate plan already satisfies physical driving constraints. Finally, although dynamical early exit comparison is computationally efficient, performing it at all decoder layers would impose unnecessary overhead. Thus, we introduce a \emph{Multi-Hop Exit Controller} that adaptively selects which layers to evaluate based on dissimilarity trends, making large jumps when predictions are far from threshold and fine-grained checks when nearing convergence.

DeeAD requires no architectural modification, no retraining, and integrates seamlessly with existing VLA models such as ORION. Through dynamic monitoring of action-space convergence, it transforms full-depth decoding into an efficient, and safety-consistent   planning process suitable for real-time autonomous driving.
\vspace{5pt}
\noindent\textbf{Contributions.}
This paper makes the following key contributions:

\begin{itemize}
\item We present a \textbf{physically grounded dynamic early-exit mechanism} for VLA planning, which terminates inference once intermediate trajectories fall within an acceptable spatial deviation from a navigation prior—enabling training-free, interpretable, and action-aligned acceleration.

\item We design two lightweight components to support efficient early termination:
(i) a \textbf{dissimilarity estimator} that measures spatial alignment between predicted and reference trajectories at negligible cost, and
(ii) a \textbf{multi-hop exit controller} that reduces redundant per-layer evaluations by adaptively skipping layers based on deviation magnitude, greatly lowering latency overhead.

\item We integrate DeeAD into the ORION VLA framework and evaluate it on Bench2Drive, achieving up to \textbf{28\% transformer-layer sparsity} and \textbf{29\% latency reduction}.
\end{itemize}

\section{Preliminaries}

\subsection{Vision-Language-Action Models for Driving}

End-to-end (E2E) autonomous driving has evolved from modular pipelines into unified learning-based policies. Traditional stacks separate perception, prediction, and planning~\cite{chauffeurnet, autopilot2020}, limiting global optimization.
Recent E2E models like UniAD~\cite{unidad2023} and VAD~\cite{jiang2023vad} predict trajectories directly from vision and route inputs using transformer backbones. While effective, these approaches are vision-only and rely on task-specific supervision.
Further advances like DriveTransformer~\cite{drivetransformer} and ORION~\cite{orion} support joint spatial-temporal reasoning. ORION introduces the Vision-Language-Action VLA pipeline, integrating image, map, and instruction for unified closed-loop planning.
Meanwhile, pre-trained VLMs developed for captioning and VQA~\cite{li2023blip, alayrac2022flamingo} are increasingly used in driving. DriveVLM~\cite{tian2024drivevlm} and EMMA~\cite{hwang2024emma} enhance grounding and scene understanding via language-conditioned perception.
These trends converge in VLA frameworks that fuse multimodal inputs to reason about driving intent and produce physically valid trajectories.
\begin{figure}[t]
  \centering
  \includegraphics[width=0.5\textwidth,trim=0cm 6cm 7cm 0cm, clip]{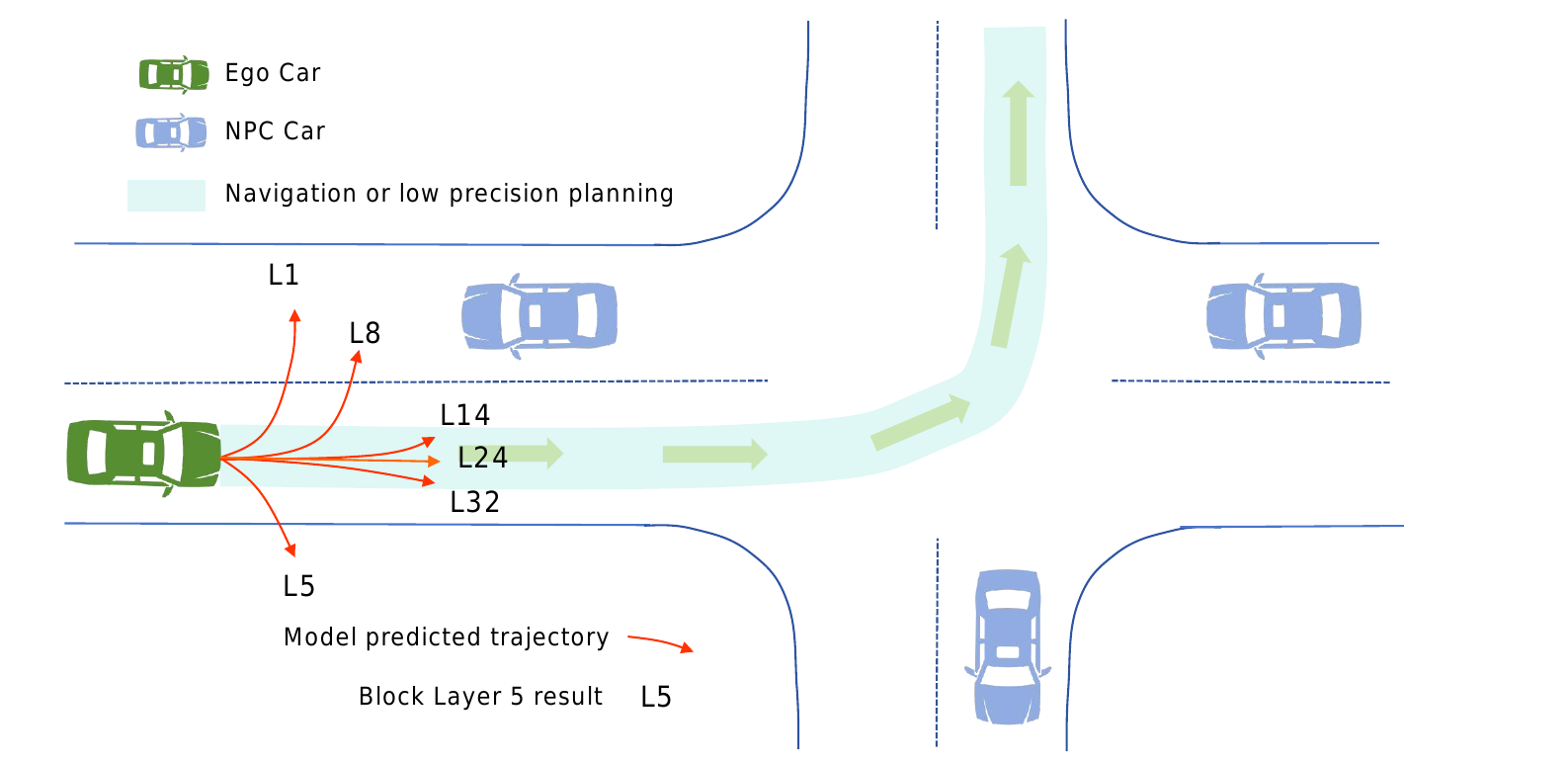}
  \vspace{-10pt}
  \caption{
Illustration of action-space early exiting. Intermediate layers (e.g., L1, L5, L8) produce inaccurate or unsafe trajectories. In contrast, deeper layers (e.g., L14, L24, L32) yield trajectories that align closely with the navigation intent. Since L14 already falls within a reasonable driving corridor, inference can safely terminate early at this point, significantly reducing computational cost without compromising driving quality.
}
  \vspace{-10pt} 
  \label{fig:traget}
\end{figure}
\subsection{Inference Bottlenecks and Dynamic Sparsity}

Current VLA models often require more than 30 transformer layers to generate control trajectories, leading to high inference latency and GPU load that challenge real-time deployment on edge devices. Existing early-exit (EE) strategies such as DeeBERT~\cite{deebert2020}, FastBERT~\cite{fastbert}, and LayerSkip~\cite{layerskip2024} accelerate inference via confidence-based termination, but they typically require retraining~\cite{cebert2024} and are sensitive to domain shifts.

More importantly, conventional EE methods ignore the physical semantics of motion. In autonomous driving, multiple trajectories can be equally feasible and safe. As long as the predicted path stays within a small tolerance tube around a safe reference, the driving behavior is acceptable. Evaluation protocols in motion forecasting benchmarks such as the Waymo Open Motion Challenge explicitly adopt this notion: a multi-modal prediction is considered successful if at least one mode lies within 2\,m of the ground-truth trajectory~\cite{waymo,modeseq2025}. This suggests that early-exit decisions need not chase the numerically optimal trajectory at the last layer; instead, they can stop once the action already lies inside a physically valid corridor.

This tolerance leads to a form of \emph{dynamic sparsity} in depth: many late transformer layers contribute only marginal refinements to the trajectory, while early and mid-depth layers already produce actions that are physically close to the final solution. Understanding how this redundancy manifests along depth is key to designing an effective early-exit policy.

\subsection{Motivation}

\begin{figure}[t]
  \centering
  \includegraphics[width=0.4\textwidth,trim=5cm 6cm 8cm 0cm, clip]{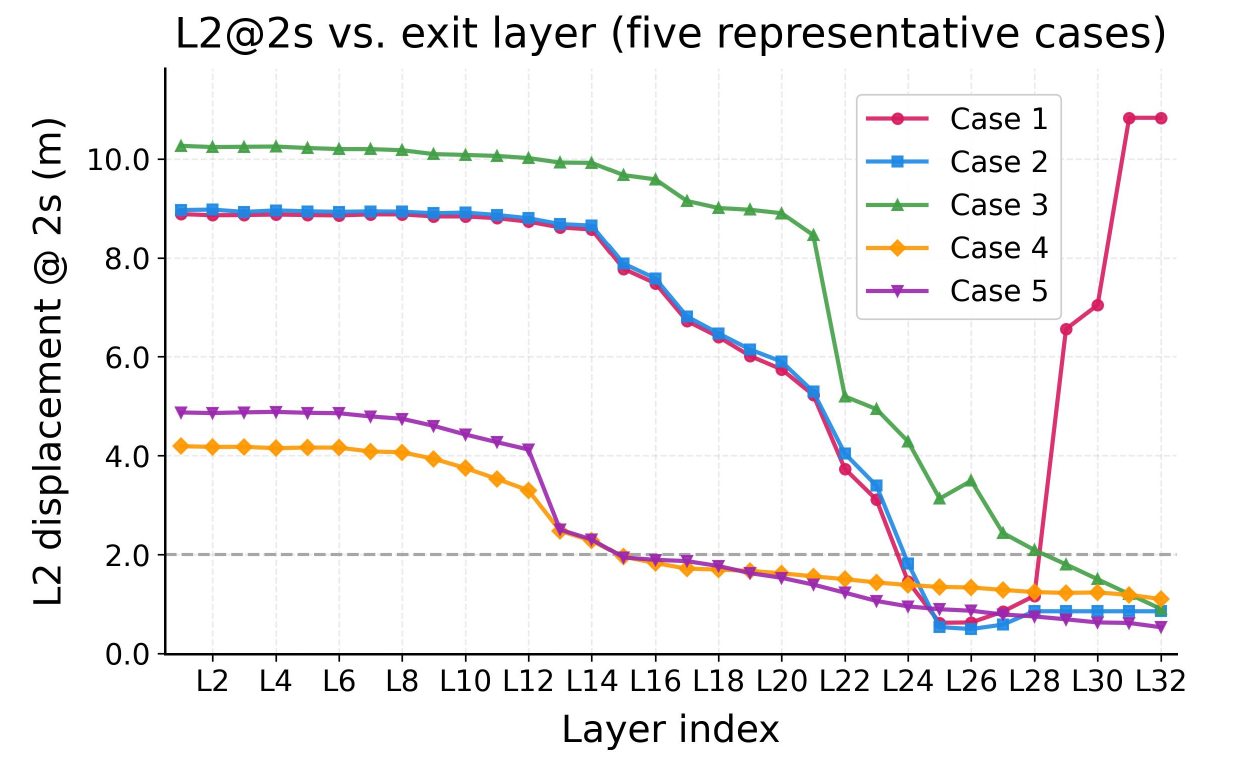}
  \vspace{-5pt}
  \caption{
  Layer-wise $\mathrm{L2}$ distance at 2s between intermediate and final trajectories on \textbf{Bench2Drive} using ORION. Each curve represents one representative case from 1,000 trials. The gray dashed line indicates the 2m tolerance threshold.
  }
  \vspace{-10pt} 
  \label{fig:l2}
\end{figure}

To characterize the depth-wise convergence behavior of trajectory decoding, we evaluate the vanilla ORION planner on 1,000 Bench2Drive scenarios, decoding intermediate trajectories after each decoder layer. For each layer, we compute the $\mathrm{L2}$ displacement at the 2-second horizon between the intermediate output and the final-layer trajectory. Figure~\ref{fig:l2} illustrates five representative examples from this analysis, with the gray dashed line indicating the 2-meter tolerance commonly adopted in motion forecasting benchmarks.

Our analysis reveals two key observations. First, the $\mathrm{L2}$ distance generally decreases gradually rather than abruptly across layers. Once the model enters an acceptable regime, adjacent layers typically differ by only several tenths of a meter. Consequently, a substantial fraction of intermediate trajectories fall below the 2m threshold well before the final layer, indicating that multiple intermediate layers can produce physically plausible trajectories.
Second, we observe that deeper layers do not universally guarantee improvement. In Case 1, for instance, the $\mathrm{L2}$ distance increases significantly after layer 25, ultimately exceeding 10 meters in the final output. This counterexample demonstrates that exhaustive layer execution can sometimes degrade planning quality.

To systematically quantify these phenomena, we extend our analysis to 640 Bench2Drive cases, identifying for each case the earliest decoder layer where the $\mathrm{L2}$ distance to the final trajectory drops below 2.0m. Table~\ref{tab:layer_l2_stat} summarizes the distribution of these valid exit points across layers.
The distribution reveals that almost no case can exit before L13, indicating that very shallow layers are insufficient for stable planning. 
Most valid exits cluster around two regions: a mid-depth band (roughly L13--L16) and the final block of layers (L24--L32).  This finding directly motivates the development of adaptive early-exit strategies that can leverage such intermediate satisfactory solutions.

Combined with the smooth error reduction observed between adjacent layers (typically several tenths of a meter near the threshold), these results suggest that dense evaluation at every layer is unnecessary. Instead, we can exploit the predictable convergence pattern to design an efficient early-exit mechanism. 

\begin{figure*}[!t] \centering \includegraphics[width=1\textwidth, height=0.4\textheight]{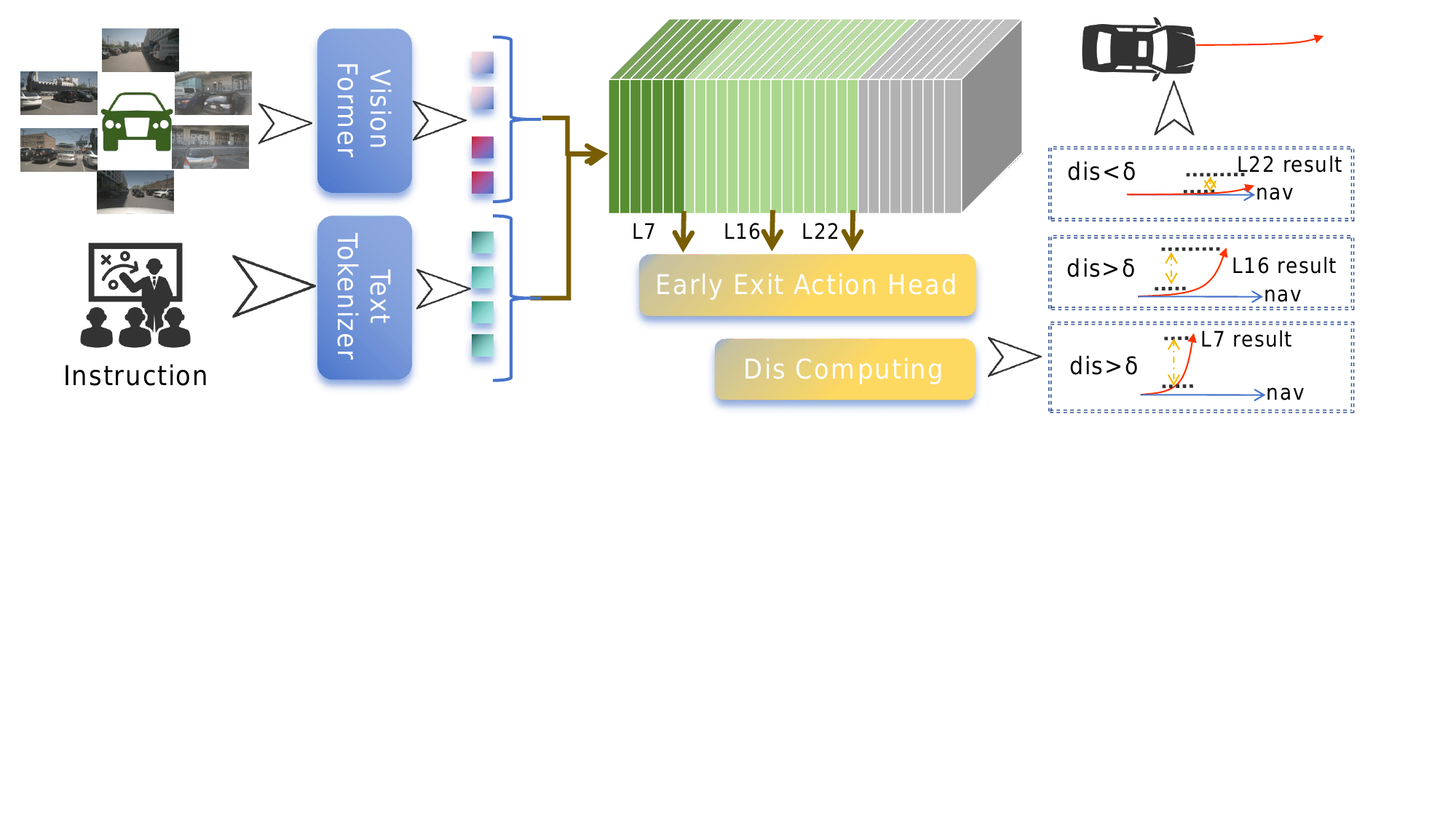} \vspace{-130pt} \caption{Overview of the proposed block-layer selective loading framework for multi-modal tasks. Inputs from the vision and text encoders are processed through a shared Transformer backbone, where task-specific layers are dynamically loaded from storage to GPU memory based on the selected cutting range.} \label{fig:overview} \end{figure*}

\section{Methodology}

We present \textbf{DeeAD}, a physically grounded early-exit framework that dynamically halts inference in VLA models once an actionable trajectory emerges. As illustrated in Fig.~\ref{fig:overview}, DeeAD integrates seamlessly into a unified Transformer-based architecture that encodes visual scenes and language instructions through a shared backbone. Rather than processing all transformer layers to completion, our system selectively computes trajectory outputs at designated checkpoints (e.g., L7, L16, L22) and compares them to a lightweight navigation prior (e.g., coarse global planner or waypoints). Once the predicted trajectory deviates within an acceptable tolerance (e.g., $<$ 2 meters), inference is terminated early, and the corresponding output is adopted for downstream control.
\begin{table}[t]
\centering
\caption{
Layer-wise distribution of early-exitable cases on Bench2Drive.
\textit{Count} is the number of cases whose earliest layer with $\mathrm{L2}<2.0$\,m is $L$.
}
\label{tab:layer_l2_stat}
\renewcommand{\arraystretch}{0.8}
\setlength{\tabcolsep}{4pt}
\resizebox{\linewidth}{!}{
\begin{tabular}{c|cccccccc}
\toprule
Layer $L$   & 1 & 2 & 3 & 4 & 5 & 6 & 7 & 8 \\
Count      & 0 & 0 & 0 & 0 & 0 & 0 & 1 & 0 \\
Percentage & 0.0\% & 0.0\% & 0.0\% & 0.0\% & 0.0\% & 0.0\% & 0.2\% & 0.0\% \\
\midrule
Layer $L$   & 9 & 10 & 11 & 12 & 13 & 14 & 15 & 16 \\
Count      & 1 & 0  & 0  & 0 & 26 & 28 & 18 & 16 \\
Percentage & 0.2\% & 0.0\% & 0.0\% & 0.0\% & 4.1\% & 4.4\% & 2.8\% & 2.5\% \\
\midrule
Layer  $L$  & 17 & 18 & 19 & 20 & 21 & 22 & 23 & 24 \\
Count      & 3  & 5  & 8  & 4  & 8  & 6  & 8  & 43 \\
Percentage & 0.5\% & 0.8\% & 1.2\% & 0.6\% & 1.2\% & 0.9\% & 1.2\% & 6.7\% \\
\midrule
Layer  $L$  & 25  & 26 & 27 & 28 & 29 & 30 & 31 & 32 \\
Count      & 146 & 1  & 26 & 32 & 26 & 3  & 5  & 226 \\
Percentage & 22.8\% & 0.2\% & 4.1\% & 5.0\% & 4.1\% & 0.5\% & 0.8\% & 35.3\% \\
\bottomrule
\end{tabular}
}
\end{table}
\subsection{Early Exit Action Head}
To enable action-level inference termination, we introduce the \textit{Early Exit Action Head} at selected transformer blocks. Unlike conventional classification or language heads, this module directly decodes intermediate representations into physically executable plans, aligning with the trajectory prediction nature of VLA models.

Formally, let $\mathcal{F}$ denote the transformer backbone with $L$ decoder layers. We define a candidate set of early-exit layers $\mathcal{L}_{\text{exit}} = \{l_1, l_2, \ldots, l_k\}$, typically concentrated in the middle and final stages (e.g., $l_i \in [12, 32]$). At each layer $l_i$, the decoder produces a hidden representation $\mathbf{h}^{(l_i)} \in \mathbb{R}^{T \times d}$, where $T$ is the planning horizon and $d$ is the embedding size. The Early Exit Action Head then predicts a 2D trajectory:
\begin{equation}
    \hat{\mathbf{P}}^{(l_i)} = \mathcal{H}^{(l_i)}\left( \mathbf{h}^{(l_i)} \right) \in \mathbb{R}^{T \times 2},
\end{equation}
where $\hat{\mathbf{P}}^{(l_i)} = \{(x_1, y_1), \ldots, (x_T, y_T)\}$ represents the predicted positions over time.

To reduce computational burden, $\mathcal{H}^{(l_i)}$ shares the same architecture as the final head and optionally reuses weights across layers. This allows intermediate outputs to be assessed efficiently without introducing significant parameter or memory overhead.

In practice, these early-exit heads are only activated when inference checks are triggered, ensuring minimal latency impact while enabling physically meaningful termination at runtime.
\subsection{Dissimilarity Estimation}

At each candidate early-exit layer $l_i \in \mathcal{L}_{\text{exit}}$, DeeAD performs an action-space comparison between the predicted trajectory and a lightweight planning prior. This evaluation determines whether the current prediction is sufficiently aligned with the navigation intent, enabling early termination.

In our implementation, we define the predicted trajectory from layer $l_i$ as $\hat{\mathbf{P}}^{(l_i)} = \{(x_t, y_t)\}_{t=1}^{T}$, and the reference trajectory—obtained from CARLA navigation waypoints or Autoware.Universe low-resolution plans—as $\mathbf{P}^{\text{ref}} = \{(x_t^{\text{ref}}, y_t^{\text{ref}})\}_{t=1}^{T}$. The spatial dissimilarity is then computed via the standard L2 metric:
\begin{equation}
    \text{Dis}^{(l_i)} = \frac{1}{T} \sum_{t=1}^{T} \left\| (x_t, y_t) - (x_t^{\text{ref}}, y_t^{\text{ref}}) \right\|_2,
\end{equation}
where $\text{Dis}^{(l_i)}$ reflects the geometric deviation from the reference path.

Early-exit is triggered once this deviation falls within a tolerance threshold $\delta$:
\begin{equation}
    \text{Exit if} \quad \text{Dis}^{(l_i)} < \delta.
\end{equation}

This L2-based formulation is simple, interpretable, and efficient ($\approx 0.2\,\text{ms per evaluation}$), making it ideal for real-time planning.

\textbf{Extension.}
Although our current experiments focus on spatial trajectories, the proposed framework is not limited to $(x,y)$ coordinates. In practical deployments, vehicle control signals—such as throttle position, brake curves, or acceleration profiles—can also be used to derive implicit motion trajectories via forward simulation or control integration. These signals can be converted to $(x,y)$ paths using kinematic models (e.g., bicycle model or PID-controlled rollout), and subsequently evaluated under the same deviation metric.

This flexibility allows DeeAD to generalize beyond coordinate-level plans and support diverse autonomous driving stacks that operate in either control or trajectory space.
\begin{table*}[!htbp]
\centering
\caption{
Main experimental results on Bench2Drive open-loop evaluation.
L2 (m) measures trajectory displacement error; Collision (\%) measures predicted trajectory overlaps. 
Sps = sparsity (\% skipped layers); Lat = per-frame inference latency (ms). 
Condition: NC = navigation command. 
Modality: C = camera.
\vspace{2pt}}
\label{tab:bench2drive_openloop_final}
\renewcommand{\arraystretch}{1.05}
\resizebox{\textwidth}{!}{%
\begin{tabular}{l|c|c|cccc|cccc|c|c|c}
\hline
\multirow{2}{*}{Method} 
& \multirow{2}{*}{Sps (\%)$\uparrow$} 
& \multirow{2}{*}{Lat (ms)$\downarrow$} 
& \multicolumn{4}{c|}{L2 (m)$\downarrow$} 
& \multicolumn{4}{c|}{Collision (\%)$\downarrow$} 
& \multirow{2}{*}{Modality} 
& \multirow{2}{*}{Condition} 
& \multirow{2}{*}{Reference} \\
\cline{4-11}
& & & 1s & 2s & 3s & Avg. & 1s & 2s & 3s & Avg. & & & \\
\hline
UniAD-Tiny~\cite{unidad2023} & 0.0 & 185 & 0.32 & 0.80 & 1.28 & 0.80 & 0.15 & 0.56 & 0.96 & 0.55 & C & NC & CVPR 23 \\
UniAD-Base~\cite{unidad2023} & 0.0 & 210 & 0.29 & 0.73 & 1.17 & 0.73 & 0.11 & 0.55 & 0.87 & 0.51 & C & NC & CVPR 23 \\
VAD~\cite{jiang2023vad}            & 0.0 & 124 & 0.36 & 0.91 & 1.46 & 0.91 & 0.13 & 0.78 & 1.11 & 0.67 & C & NC & ICCV 23 \\
\hline
ORION~\cite{orion}                & 0.0  & 381 & 0.26 & 0.63 & 1.12 & 0.67 & 0.21 & 0.46 & 0.74 & 0.47 & C & NC & ICCV 25 \\

ORION + DeeAD (0.5 m)           & 21.3 & 322 & \textbf{0.24} & \textbf{0.57} & \textbf{0.92} & \textbf{0.58} & 0.18 & 0.41 & 0.68 & \textbf{0.42} & C & NC & -- \\
ORION + DeeAD (1.0 m)           & 24.8 & 311 & 0.62 & 0.98 & 1.21 & 0.93 & 0.20 & 0.43 & 0.69 & 0.44 & C & NC & -- \\
ORION + Fixed-EE & 25.0 & 305 & 1.12 & 1.67 & 2.10 & 1.63 & 0.90 & 1.27 & 1.82 & 1.32 & C & NC & -\\
ORION + DeeAD (2.0 m)           & \textbf{28.0} & \textbf{270} & 0.73 & 1.12 & 1.54 & 1.13 & 0.21 & 0.45 & 0.70 & 0.45 & C & NC & -- \\
\hline
\end{tabular}
}
\end{table*}

\begin{figure*}[t]
\centering
\includegraphics[width=1\textwidth,,trim=0cm 12.5cm 0cm 0cm, clip]{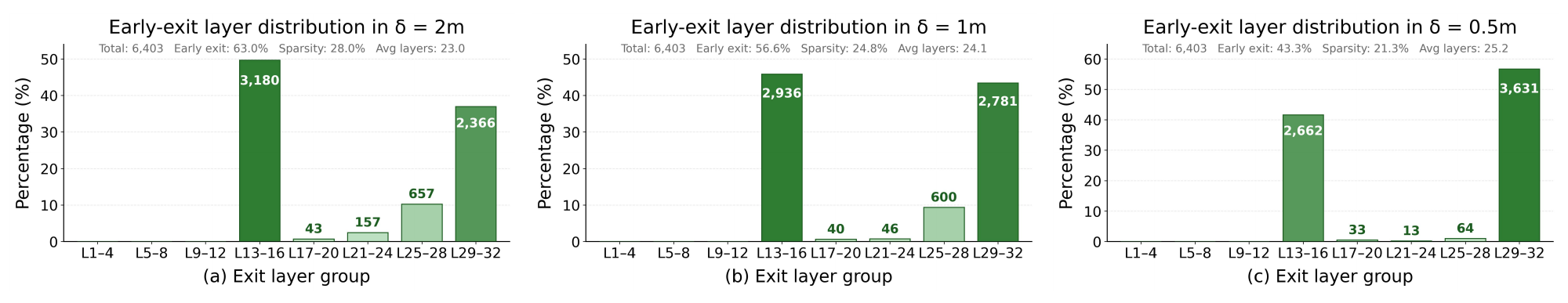}
\vspace{-20px}
\caption{Early-exit layer distributions under different spatial tolerance thresholds $\delta$. As $\delta$ decreases, the model becomes more conservative, shifting exits toward deeper layers.}
\label{fig:distribute}
\end{figure*}
\subsection{Multi-Hop Exit Controller}

To reduce redundant computations in early-exit checking, DeeAD adopts a deterministic \textit{multi-hop exit controller} that adaptively skips layers according to the current dissimilarity score.

From the analysis in Fig.~\ref{fig:l2} and Table~\ref{tab:layer_l2_stat}, two empirical facts are observed:
\begin{itemize}
    \item Valid early-exitable trajectories almost never appear before layer 13; the first layers mainly perform coarse stabilization.
    \item For the five representative cases in Fig.~\ref{fig:l2}, the layer-wise $\mathrm{L2}$ difference after L13 is usually within $0$–$1.6$\,m (with the 5th--95th percentile roughly $0.04$–$1.3$\,m) once rare re-routing spikes are excluded.
\end{itemize}
This means that, in the ``normal'' regime, each additional layer typically reduces the L2 distance by a fraction of a meter up to about 1--1.5\,m.  
Consequently, when the current trajectory is far away from the tolerance tube, several layers are required before it can plausibly enter the safe region.

Motivated by this, we design a simple rule-based controller whose skip stride is proportional to how far the current prediction is from the tolerance threshold~$\delta$.  
Let $\text{Dis}^{(l)}$ denote the dissimilarity score (e.g., L2@2s) at layer $l$.  
The controller sets the stride $s$ as:
\begin{equation}
s =
\begin{cases}
8, & \text{if } \text{Dis}^{(l)} > 8*\delta, \\
4, & \text{if } \text{Dis}^{(l)} > 4*\delta, \\
2, & \text{if } \text{Dis}^{(l)} > 2*\delta, \\
1, & \text{otherwise.}
\end{cases}
\end{equation}
The next exit check is then performed at layer $l + s$.

This design reflects a coarse-to-fine search along depth.  
When $\text{Dis}^{(l)}$ is much larger than $\delta$ (e.g., more than $8*\delta$), the statistics above indicate that even the maximal per-layer improvement cannot close the gap within just a few layers, so an aggressive jump ($s=8$) is safe and efficient.  
As the score approaches the tolerance band (between $2*\delta$ and $4*\delta$), the controller automatically switches to medium or small hops ($s=2$ or $1$), ensuring that near-feasible trajectories are examined more closely and that we do not skip over the admissible exit range.

In practice, this multi-hop controller substantially reduces the number of layers at which exit checks are evaluated, while still respecting the layer-wise convergence patterns observed on Bench2Drive and preserving all valid early-exit opportunities.

\section{Experiments}
\subsection{Experimental Setup}

\textbf{Dataset.}  
We conduct all experiments on the \textbf{Bench2Drive} benchmark~\cite{bench2drive2024}, a closed-loop evaluation suite derived from the CARLA simulator. It offers over 1,000 driving clips for training and 220 diverse route segments for validation. These routes feature challenging urban layouts including roundabouts, unprotected left turns, occluded merges, and multi-agent intersections. Each frame contains synchronized RGB views from five cameras, high-level navigation commands, and ground-truth trajectory rollouts sampled at 2\,Hz. 

\textbf{Model.}  
We adopt the official ORION model as our base VLA architecture. It consists of a QT-Former backbone, a BEV encoder with language fusion, and a trajectory generator. DeeAD is injected into the planning head as a runtime wrapper with no architectural changes or retraining. Action Heads are shared across layers and reused from ORION's original decoder.

\textbf{Metrics.}  
We evaluate four key aspects:
\begin{itemize}
    \item \textbf{L2@t (m)}: Open-loop trajectory displacement at 1s, 2s, and 3s horizon;
    \item \textbf{Collision rate (\%)}: Whether the predicted path intersects with known objects (per simulation);
    \item \textbf{Latency (ms)}: Mean forward time per sample, excluding sensor pre-processing;
    \item \textbf{Sparsity (\%)}: Proportion of skipped transformer layers due to early exit.
\end{itemize}

These metrics reflect short-term accuracy, safety, computational efficiency, and dynamic sparsity, respectively.

\textbf{Platform.}  
All experiments are conducted on a workstation running Ubuntu 22.04, equipped with dual Intel Xeon Silver 4314 CPUs (32 cores), 512\,GB RAM, and 2$\times$NVIDIA L20 GPUs (40\,GB each). We use PyTorch 1.8 with CUDA 11.8.

\subsection{Main Results}

Table~\ref{tab:bench2drive_openloop_final} summarizes the open-loop results on Bench2Drive. 
Among existing planners, ORION achieves the best overall trade-off between accuracy and safety, but incurs the highest latency (381\,ms). 
Integrating DeeAD into ORION produces a family of operating points controlled by the tolerance $\delta$, tracing out a good balance between trajectory quality and computational cost.

With a strict tolerance of $\delta=0.5$\,m, DeeAD already yields clear benefits: it skips 21.3\% of layers on average and reduces latency from 381\,ms to 322\,ms (about 15\% reduction), \emph{while also improving} trajectory quality and safety. 
Compared to vanilla ORION, the average L2 error drops from 0.67\,m to 0.58\,m and the average collision rate decreases from 0.47\% to 0.42\%. 
This indicates that some late layers in ORION over-refine the trajectory, and action-guided early exit can both accelerate and stabilize planning.

At $\delta=1.0$\,m, DeeAD becomes more aggressive: it skips 24.8\% of layers and further reduces latency to 311\,ms (around 18\% reduction). 
The L2 error increases moderately (0.93\,m vs.\ 0.67\,m), but the collision rate remains comparable or slightly better (0.44\% vs.\ 0.47\%), showing that safety is largely preserved even when exiting earlier.

Pushing tolerance to $\delta=2.0$\,m leads to the fastest configuration: DeeAD skips 28.0\% of layers and achieves 270\,ms latency (roughly 29\% faster than ORION). 
This comes at the cost of a higher average L2 (1.13\,m), yet the collision rate (0.45\%) is still on par with the vanilla model, suggesting that most early exits remain physically reasonable.

Finally, DeeAD consistently outperforms the confidence-based \textit{Fixed-EE} baseline. 
Although ORION + Fixed-EE attains similar sparsity (25.0\%) and even slightly lower latency (305\,ms), it produces much worse trajectories (Avg.\ L2 = 1.63\,m) and significantly more collisions (1.32\%). 
This contrast highlights the importance of grounding early exit decisions in the action space rather than relying solely on confidence scores.

\subsection{Exit Layer Distribution}
\begin{table}[!htbp]
\centering
\caption{Per-layer overhead introduced by our early-exit instrumentation (micro-benchmark, single L20 GPU).}
\label{tab:overhead_micro}
\renewcommand{\arraystretch}{1.05}
\resizebox{0.9\linewidth}{!}{%
\begin{tabular}{l|c}
\hline
Component & Cost (ms / layer) \\
\hline
Planning metric evaluation ($\mathrm{L2}(\hat{\mathbf{p}}^{(l)}, \mathbf{p}_{ref})$) & 0.20 \\
Intermediate feature extraction (from layer $l$) & 0.70 \\
Action Head projection ($\mathbf{h}^{(l)} \!\rightarrow\! \hat{\mathbf{p}}^{(l)}$) & 4.00 \\
\hline
\textbf{Total per layer (added)} & \textbf{4.90} \\
\hline
\end{tabular}
}
\end{table}
\begin{table}[!htbp]
\centering
\caption{End-to-end latency under different early-exit depths.}
\label{tab:overhead_scenarios}
\renewcommand{\arraystretch}{1.05}
\resizebox{0.95\linewidth}{!}{%
\begin{tabular}{c|c|c|c}
\hline
Exit layer $l^\ast$ & Added EE cost (ms) & Total latency (ms) & $\Delta$ vs. baseline \\
\hline
16 & $\approx 1 \times 4.9 = 4.9$ & \textbf{203} & \textbf{-178 ms} \\
32 (early exit) & $\approx 16 \times 4.9 = 78.4$ & \textbf{440} & \textbf{+78.4 ms} \\
32 (no early exit) & 0 & \textbf{381} & \textbf{0ms} \\
\hline
\end{tabular}
}
\end{table}
Fig.~\ref{fig:distribute} illustrates the distribution of exit layers across varying tolerance thresholds ($\delta = 0.5$, $1.0$, and $2.0$\,m). 

As expected, a stricter threshold ($\delta = 0.5$\,m) pushes most exits toward the final layers: over 56.7\% of all samples exit in the final group (L29–32), indicating the need for more refined reasoning in safety-critical situations. Only 21.3\% of all layers are skipped on average in this setting.

As the tolerance relaxes, exits occur significantly earlier. Under $\delta = 1.0$\,m, 45.9\% of samples exit before L25, and the average depth drops to 24.1 layers. With the loosest constraint ($\delta = 2.0$\,m), over 63.0\% of samples exit early, with a notable 49.7\% terminating at L13–16 and L29–32. This leads to a substantial 28.0\% sparsity, reducing the average layer depth to 23.0.

These results suggest that many driving scenes, especially structured ones, can be handled without full-layer reasoning. DeeAD adapts its exit strategy based on tolerance, enabling deeper reasoning only when necessary while accelerating inference in routine scenarios.

\subsection{Runtime Overhead Analysis}

DeeAD adds only lightweight computation on top of the original planner. As summarized in Table~\ref{tab:overhead_micro}, the early-exit instrumentation introduces 4.90\,ms of overhead per evaluated layer, dominated by the Action Head projection that decodes candidate trajectories from the hidden state (4.00\,ms). In contrast, the planning metric evaluation $\mathrm{L2}(\hat{\mathbf{p}}^{(l)}, \mathbf{p}_{ref})$ and intermediate feature extraction together contribute less than 1\,ms per layer (0.20\,ms and 0.70\,ms, respectively), so most of the added cost comes from a single extra forward pass through the Action Head at each candidate exit layer.

Table~\ref{tab:overhead_scenarios} reports the end-to-end latency under different exit depths. When early exit is triggered at mid-depth (layer $l^\ast=16$), total latency drops from 381\,ms (no early exit) to 203\,ms, yielding a 47\% speed-up while incurring only one layer-level check (approximately 4.9\,ms of added cost). In the corner case where DeeAD is enabled but all samples exit at the last candidate layer ($l^\ast=32$), the added early-exit computation is still bounded: latency increases from 381\,ms to 440\,ms, i.e., about +78.4\,ms overhead, without introducing large tail-latency spikes. If early exit is turned off entirely, the system falls back to the 381\,ms baseline, showing that DeeAD can be deployed without degrading the original worst-case runtime.

\subsection{Ablation Study}

To evaluate the contribution of each design in DeeAD, we perform comprehensive ablations on the Bench2Drive benchmark. 
The full system includes: (1) \textit{action-guided early exit} via trajectory deviation, 
(2) \textit{multi-hop exit controller} that adaptively chooses the next checked layer, 
and (3) \textit{a physically grounded tolerance threshold} ($\delta = 1.0$). 
Higher Sps indicates that more layers are skipped (i.e., higher sparsity).

\begin{table}[!htbp]
\centering
\caption{Ablation study on the Bench2Drive validation set. 
Baseline = DeeAD with action-aligned exit, multi-hop controller (starting from L13), and $\delta=1.0$.}
\label{tab:ablation}
\renewcommand{\arraystretch}{1.1}
\resizebox{\linewidth}{!}{%
\begin{tabular}{l|c|c|c|c|l}
\hline
Method Variant & L2 & Collision & Lat. & Sps & Description \\
\hline
DeeAD (Full) & 0.93 & 0.44 & 311 & 24.8 & action-guided + multi-hop \\
\hline
(B1) Fixed Exit Depth & 1.17 & 1.07 & 306 & 27.5 & global fixed exit layer \\
(B2) Full Scan from L1 & 0.91 & 0.43 & 520 & 25.0 & check every layer from L1 \\
(B3) Full Scan from L13 & 0.94 & 0.46 & 366 & 21.7 & check every layer from L13 \\
(B4) Loose Threshold ($\delta=2.0$) & 1.12 & 0.60 & \textbf{277} & \textbf{30.2} & early, risky exits \\
(B5) Strict Threshold ($\delta=0.5$) & \textbf{0.56} & \textbf{0.40} & 354 & 16.3 & late, safer exits \\
\hline
\end{tabular}
}
\end{table}

\textbf{(B1) Action-Guided Exit vs Fixed Exit Depth.}  
Replacing our action-aligned mechanism with a global fixed exit depth harms planning quality.  
Although the fixed-depth variant skips slightly more layers (Sps $27.5$ vs.\ $24.8$), it cannot adapt to scene difficulty, leading to higher L2 error (+0.24\,m) and over twice the collision rate (0.44\% $\rightarrow$ 1.07\%).

\textbf{(B2) Full Scan from Layer 1.}  
Starting exit checks from the first layer and evaluating every subsequent layer yields a marginal L2 gain (0.93\,m $\rightarrow$ 0.91\,m) but dramatically increases latency (311\,ms $\rightarrow$ 520\,ms) with almost no change in sparsity.  
This confirms our earlier observation that almost no valid exits occur before L13, so exhaustive scanning of shallow layers is largely redundant.

\textbf{(B3) Full Scan from Layer 12 vs Multi-Hop.}  
If we still start from L13 but disable the multi-hop controller (i.e., move layer-by-layer), L2 and collision remain similar to the full method, yet latency rises to 366\,ms and sparsity drops from 24.8 to 21.7.  
Thus, even when the starting depth is reasonable, adaptive strides are important for avoiding unnecessary exit evaluations.

\textbf{(B4, B5) Effect of Tolerance Threshold $\delta$.}  
Loose tolerances (e.g., $\delta=2.0$\,m) encourage aggressive early exits, achieving the highest sparsity (30.2) and lowest latency (277\,ms) but at the cost of degraded accuracy and safety.  
Conversely, stricter tolerances (e.g., $\delta=0.5$\,m) produce the safest plans (L2=0.56\,m, lowest collision) but significantly reduce sparsity.  
Our default $\delta=1.0$ strikes a balanced trade-off between decision quality and inference efficiency.

\section{Conclusion}
We present \textbf{DeeAD}, a physically grounded early-exit framework for VLA models in autonomous driving.  DeeAD dynamically monitors the convergence of action-space predictions and terminates inference once the trajectory aligns with a lightweight planning prior. To balance accuracy and efficiency, we further propose a multi-hop controller that adaptively skips layers based on spatial deviation trends. Experiments on Bench2Drive demonstrate that DeeAD achieves significant latency reduction (up to 29\%) with minimal performance degradation. Our findings highlight that planning-aware early-exit strategies, rooted in physical feasibility, enable interpretable, efficient, and safe deployment of large-scale driving models.
\bibliographystyle{ACM-Reference-Format}
\bibliography{sample-base}

\end{document}